\documentclass[conference]{IEEEtran}
\IEEEoverridecommandlockouts
\usepackage{cite}
\usepackage{amsmath,amssymb,amsfonts}
\usepackage{algorithmic}
\usepackage{graphicx}
\usepackage{textcomp}
\usepackage{xcolor}
\usepackage{titlesec,subcaption,hyperref,enumitem}
\def\BibTeX{{\rm B\kern-.05em{\sc i\kern-.025em b}\kern-.08em
    T\kern-.1667em\lower.7ex\hbox{E}\kern-.125emX}}
\begin{document}

\IEEEsettextwidth{0.75in}{0.75in}
\IEEEsetsidemargin{i}{0.75in}
\IEEEsettextheight{0.78in}{0.75in}
\def\IEEEtitletopspace{0.27in}
\IEEEsettopmargin{t}{0.78in}

\setlength{\parskip}{0pt}
\titlespacing*{\subsubsection}
{0pt}{.4ex plus 0ex minus .2ex}{0ex plus 0ex}
% {0pt}{5.5ex plus 1ex minus .2ex}{4.3ex plus .2ex}
\titlespacing*{\subsection}
{0pt}{.6ex plus 0ex minus .3ex}{0ex plus 0ex}
\titlespacing*{\section}
{0pt}{.7ex plus 0ex minus .3ex}{0ex plus 0ex}

\title{BlanketSet - A clinical real-world in-bed action recognition and qualitative semi-synchronised MoCap dataset\\
\thanks{This work is financed by National Funds through the Portuguese funding agency, FCT - Fundação para a Ciência e a Tecnologia, within project LA/P/0063/2020 as well as under the scope of the CMU Portugal (Ref PRT/BD/152202/2021).}
}

% \author{\IEEEauthorblockN{1\textsuperscript{st} João Carmona}
% City, Cohttps://www.overleaf.com/project/6414b8ba78dd7c6799336a48untry \\
% \and
% \and
% \and
% \and
% \and

\author{\IEEEauthorblockN{João Carmona\IEEEauthorrefmark{1}\IEEEauthorrefmark{2}, Tamás Karácsony\IEEEauthorrefmark{1}\IEEEauthorrefmark{2}\IEEEauthorrefmark{3}, João Paulo Silva Cunha\IEEEauthorrefmark{1}\IEEEauthorrefmark{2} \IEEEmembership{Senior Member, IEEE}}
\IEEEauthorblockA{\IEEEauthorrefmark{1}Center for Biomedical Engineering Research, INESC TEC, Porto, Portugal}
\IEEEauthorblockA{\IEEEauthorrefmark{2}Faculty of Engineering (FEUP), University of Porto, Porto, Portugal}
\IEEEauthorblockA{\IEEEauthorrefmark{3}Robotics Institute, Carnegie Mellon University, Pittsburgh, PA, USA}
}

\maketitle

\begin{abstract}
Clinical in-bed video-based human motion analysis is a very relevant computer vision topic for several relevant biomedical applications. Nevertheless, the main public large datasets (e.g. ImageNet or 3DPW) used for deep learning approaches lack annotated examples for these clinical scenarios. To address this issue, we introduce BlanketSet, an RGB-IR-D action recognition dataset of sequences performed in a hospital bed. This dataset has the potential to help bridge the improvements attained in more general large datasets to these clinical scenarios. Information on how to access the dataset is available at \href{https://rdm.inesctec.pt/dataset/nis-2022-004}{rdm.inesctec.pt/dataset/nis-2022-004}.

% Recent advancements in computer vision have drastically improved human motion analysis in videos. However, these improvements have not yet fully carried over into clinical in-bed scenarios due to the lack of public datasets representative of this scenario. To address this issue, we introduce BlanketSet, an RGB-IR-D action recognition dataset of sequences performed in a hospital bed. This dataset has the potential to help bridge the improvements attained in general use cases to these clinical scenarios. Information on how to access the dataset is available at \href{https://rdm.inesctec.pt/dataset/nis-2022-004}{rdm.inesctec.pt/dataset/nis-2022-004}.
\end{abstract}


\section{Introduction}
\label{sec:intro}
Human motion analysis in videos has recently seen drastic improvements due to advances in Deep Learning (DL). However, almost all the research in this area was focused on the most general context of people standing up without any occlusions. 
This, paired with the difficulty of generalizing DL models beyond the domains of their training datasets, means that human motion analysis for patients in bed is sorely lacking when compared with more common scenarios.

In order to bridge this domain gap, we introduce BlanketSet, an action recognition dataset of people in a hospital bed which consists of 405 RGB-IR-D recordings of 14 participants performing 8 different movement sequences. 

Each recording was repeated 3 times with different levels of blanket occlusion, so it can be used not just for action recognition, but also for qualitative evaluation of human motion analysis systems. Even though there are no ground truth poses or body shapes, the actions are carried out in a semi-synchronized manner, as described in (\ref{sequences}). As an example of this second use case, in this work we used BlanketSet to evaluate the real-world performance of the pipeline implemented in \cite{blanketgen} and found that BlanketGen improved the performance of a DL human pose estimation system with statistical significance when full-body blanket occlusions were present in the real world.

\section{Related work}
In-bed movement monitoring is done in different contexts, the most relevant ones being epileptic seizure analysis and sleep analysis.

In the area of automatic epileptic seizure classification, there has been significant research, in \cite{neurokinect} and \cite{deepepil} a system was set up to record RGB-IR-D videos synchronized with electroencephalography data of patients staying at the University of Munich Epilepsy Monitoring Unit (EMU) and then have them manually labeled by clinicians. IR data acquired with that system was used in \cite{karacsonyClassification2020} and \cite{karacsonyNature} to explore deep learning action recognition at EMUs. Data recorded with this system is very promising for future research in this area but is not publicly available, which limits its accessibility.

In the area of sleep analysis, the use of data collected from cameras positioned above the bed has been explored as a low-cost non-invasive approach with promising results \cite{Grimm2016, Yang2017, Metsis2014}. However, the only image dataset of people lying in beds that is publicly available is SLP \cite{slp}, which contains images of people in static positions along with 2D position ground truth; it is an useful resource but lacks the temporal dimension that is very relevant in contexts with considerable movement such as the analysis of epileptic seizures.

Image and video-based DL systems have been shown to work exceptionally well in tasks related to human motion recognition and classification in general use cases \cite{hybrik, dynaboa, deciwatch}, therefore it is reasonable to expect that they would also be able to perform well in the more specific use cases of sleep analysis and epileptic seizure analysis. As with everything related to deep learning, however, large amounts of data are required and, due to the lack of publicly available datasets in these areas, each separate research effort has also had to include the acquisition of its own dataset which significantly hampers the efficiency of the research. The acquisition and publication of datasets such as \cite{3dhp, 3DPW, h36m, kinetics, imagenet_cvpr09} were instrumental to the incredible results attained in the more general use cases.

There have also been research efforts into utilizing these broader datasets in epileptic seizure analysis: \cite{karacsonyClassification2020} employed transfer learning from the Kinetics-400 \cite{kinetics} and ImageNet \cite{imagenet_cvpr09} datasets to discriminate between two classes of epileptic seizures, and BlanketGen \cite{blanketgen} augmented the 3DPW dataset \cite{3DPW} with synthetic blanket occlusions to improve the performance of HybrIK \cite{hybrik} in these scenarios.

\section{Methods}
\subsection{Data Acquisition}
The dataset was acquired with an Azure Kinect at the Epilepsy Monitoring Unit (EMU) of the University Hospital Center of São João during multiple sessions over the course of several months. Covid restrictions changed while the dataset was acquired, this led to the participants wearing masks in some recordings, but not in others. One person participated both before and after the restrictions changed, and therefore both with and without a mask. 

To each participant, the data to be recorded was explained, as well as their privacy rights, before they each signed an informed consent form in which they agreed to the recording and publication of their data in the dataset.

\subsection{Dataset Design and Parameters}
In order to control for as many variables as possible within reason, the different variables of the dataset were split into controlled and semi-controlled variables.
\textbf{Controlled variables} covered the dimension space as evenly as possible.
\begin{itemize}[topsep=0.5em, partopsep=0em,itemsep=-0.2em]
    \item \textbf{Movement sequences}: 8 movement sequences were performed, which are explained in detail in section \ref{sequences}.
    \item \textbf{Blanket Initial Position (BIP)}: There were 3 different initial positions for the blanket: no blanket, fully covering the subject, and pulled down so that only about the feet and lower legs were covered.
    \item \textbf{Blanket}: Originally 3 different blankets of different thicknesses, colors, and weights were alternated between. In the recordings after the restrictions were lifted, 3 other blankets were used.
\end{itemize}

\begin{figure}[h]
    \setlength{\belowcaptionskip}{-10pt}
    \centering
    \includegraphics[width=7cm]{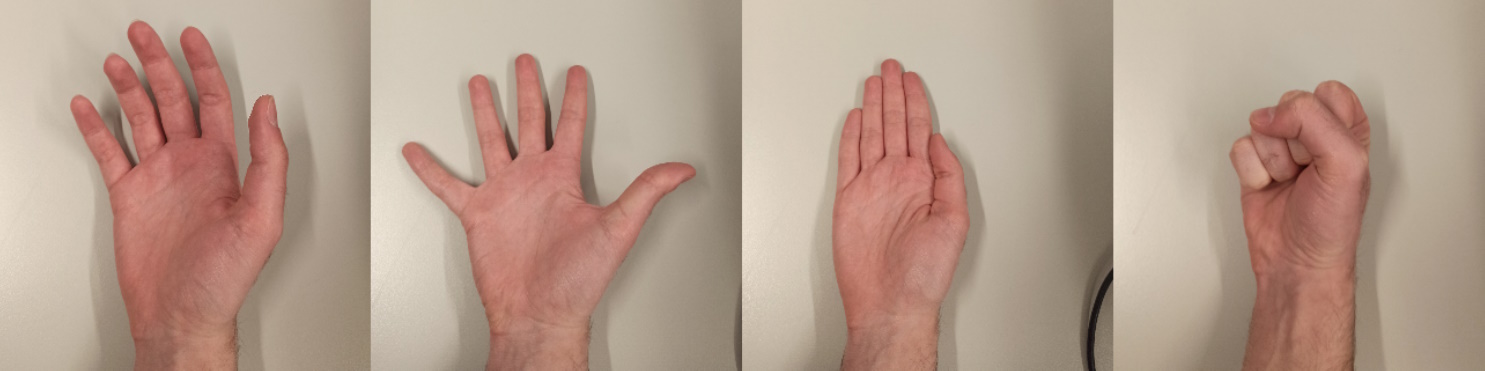}
    \caption{The 4 hand positions from left to right: relaxed, fingers spread, fingers stretched touching, and closed fist}
    \label{fig:handpos}
\end{figure}

\textbf{Semi-controlled variables} were randomly sampled from a uniform distribution so as to avoid correlations between them and other variables.
\begin{itemize}[topsep=0.5em, partopsep=0em,itemsep=-0.2em]
    \item \textbf{Lighting}: 4 different lighting setups were used: natural lightning controlled by having the window blinds up or down, and having the lights on or off.
    \item \textbf{Hand position}: Four different hand positions were included: relaxed, fingers spread, fingers stretched touching, and closed fist. Figure \ref{fig:handpos} displays these hand positions.
    % 4 hand positions were covered, they are covered in section \ref{handpos}.
    \item \textbf{Time between position changes}: 30 to 120 beats per minute (BPM) (uniformly sampled with integer precision).
\end{itemize}

\subsection{Movement Sequences} \label{sequences}
Each movement sequence consists of switching between a few positions repeatedly, each recording contains 20 position changes. The emphasis was on ensuring that the motions were repeated properly over ensuring they were done perfectly as described. In order to cover all the variations of the controlled variables, each sequence was repeated 9 times, once for every combination of BIP and blanket. The semi-controlled variables were changed with the blanket, such that for every recording with one BIP, there be recordings with the other BIPs and no other changed variables.

\begin{figure}[h!]
\begin{subfigure}{.49\linewidth}
  \centering
  \includegraphics[width=.75\linewidth]{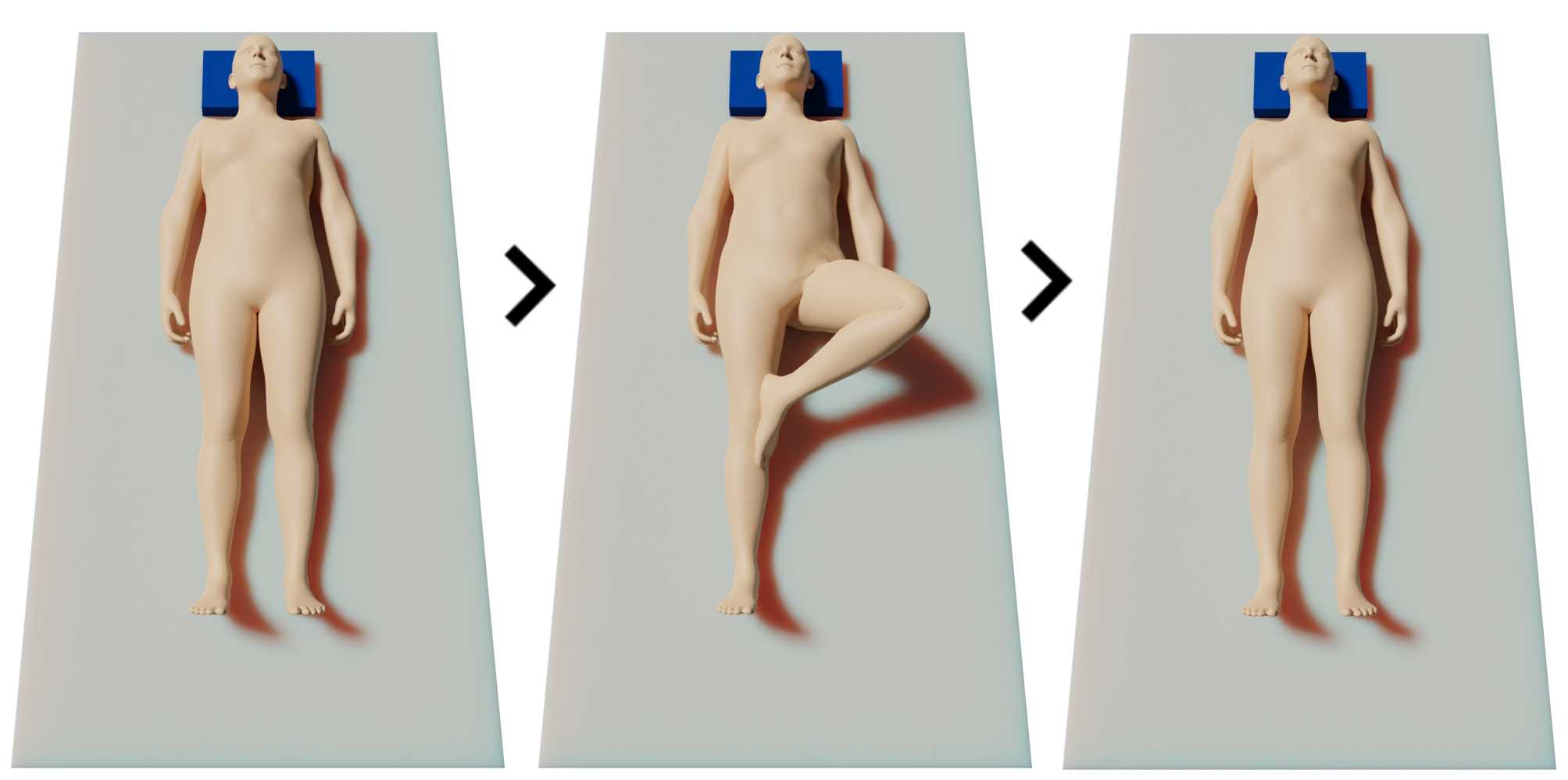}  
  \caption{Foot to knee}
  \label{fig:sub1}
\end{subfigure}
\begin{subfigure}{.49\linewidth}
  \centering
  \includegraphics[width=.75\linewidth]{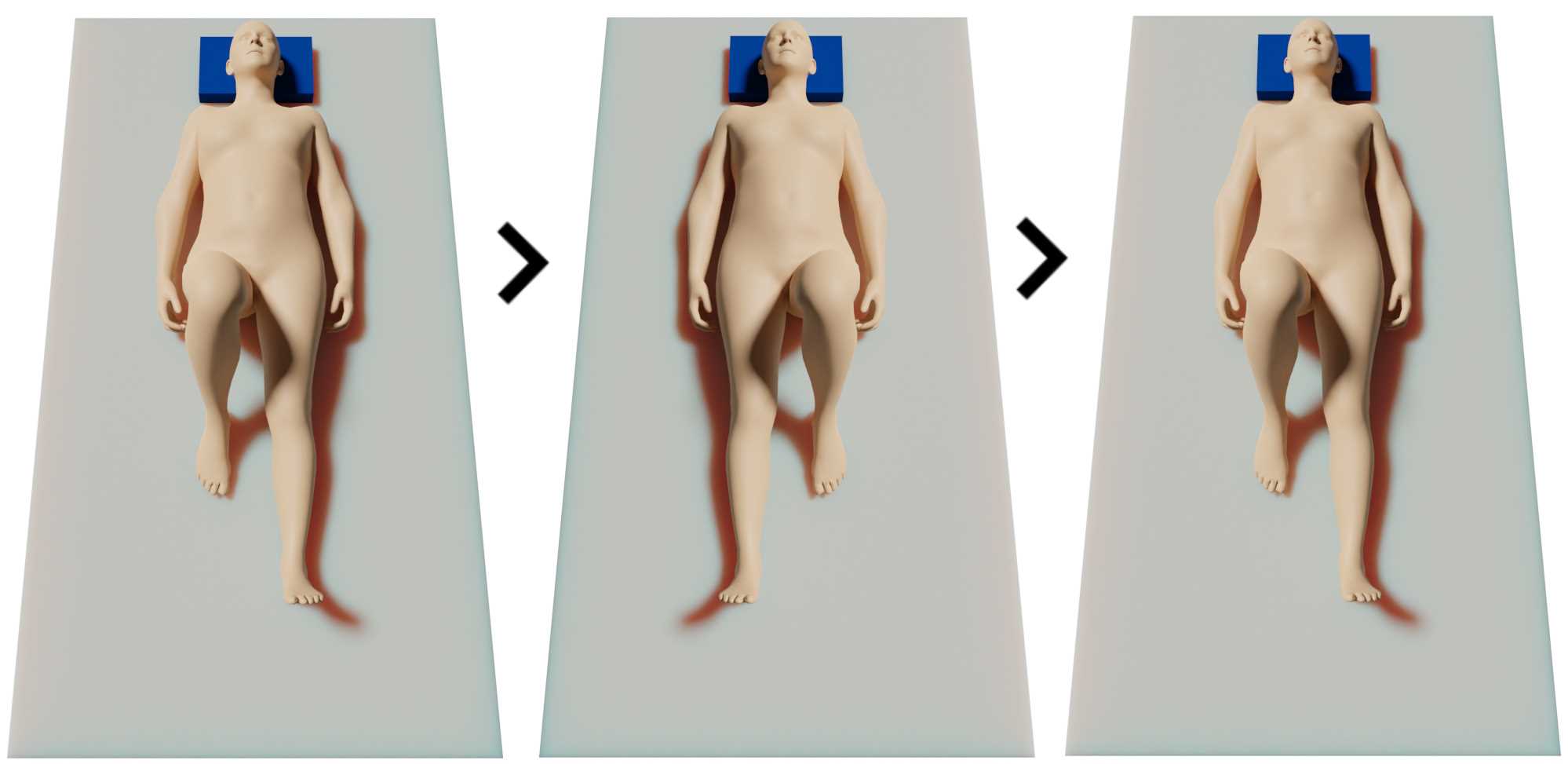}  
  \caption{Knee bend}
  \label{fig:sub2}
\end{subfigure}
\begin{subfigure}{.49\linewidth}
  \centering
  \includegraphics[width=.75\linewidth]{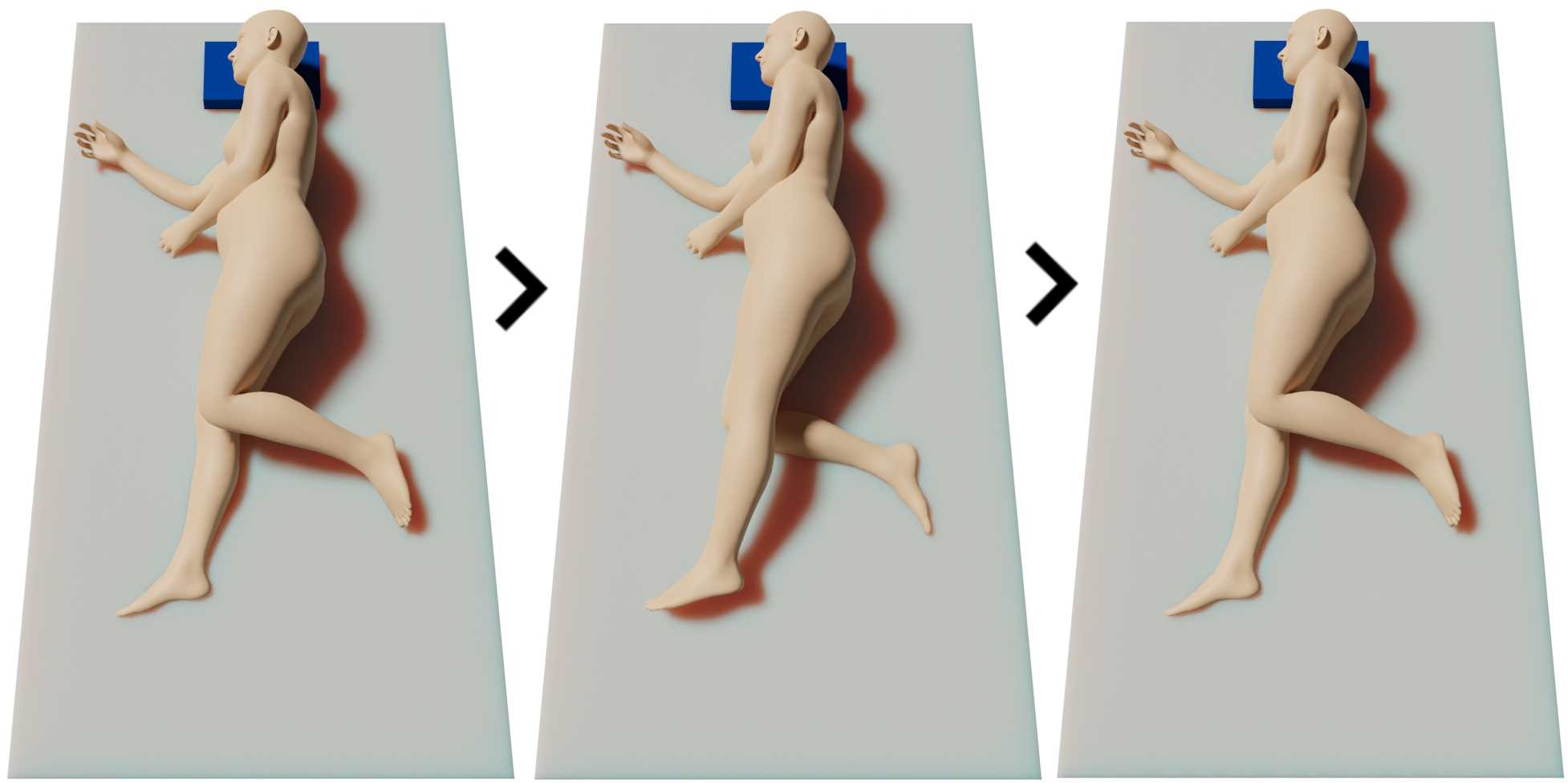}  
  \caption{Swinging legs}
  \label{fig:sub3}
\end{subfigure}
\begin{subfigure}{.49\linewidth}
  \centering
  \includegraphics[width=.75\linewidth]{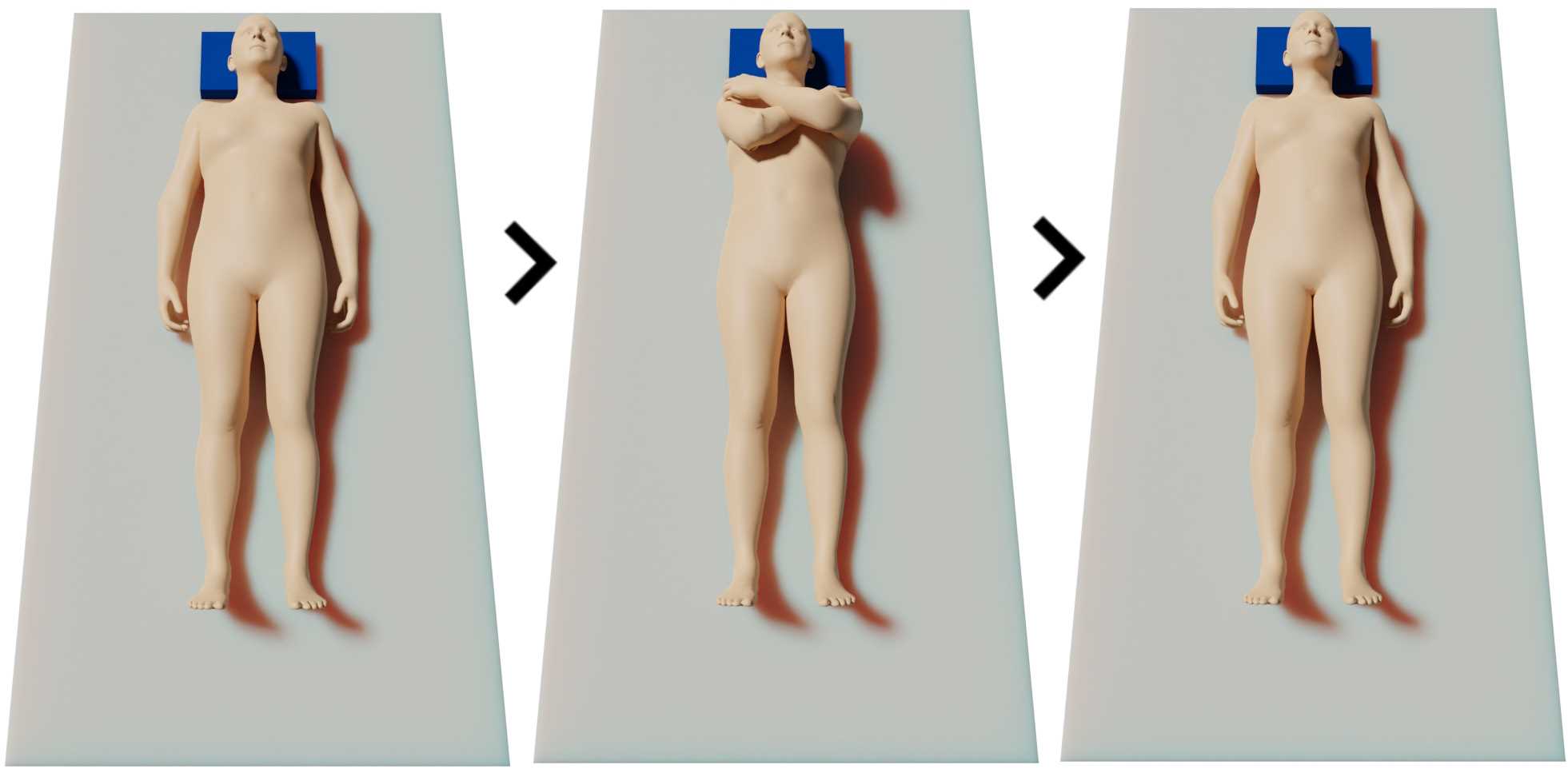}  
  \caption{Hands to shoulders}
  \label{fig:sub4}
\end{subfigure}
\begin{subfigure}{.49\linewidth}
  \centering
  \includegraphics[width=.75\linewidth]{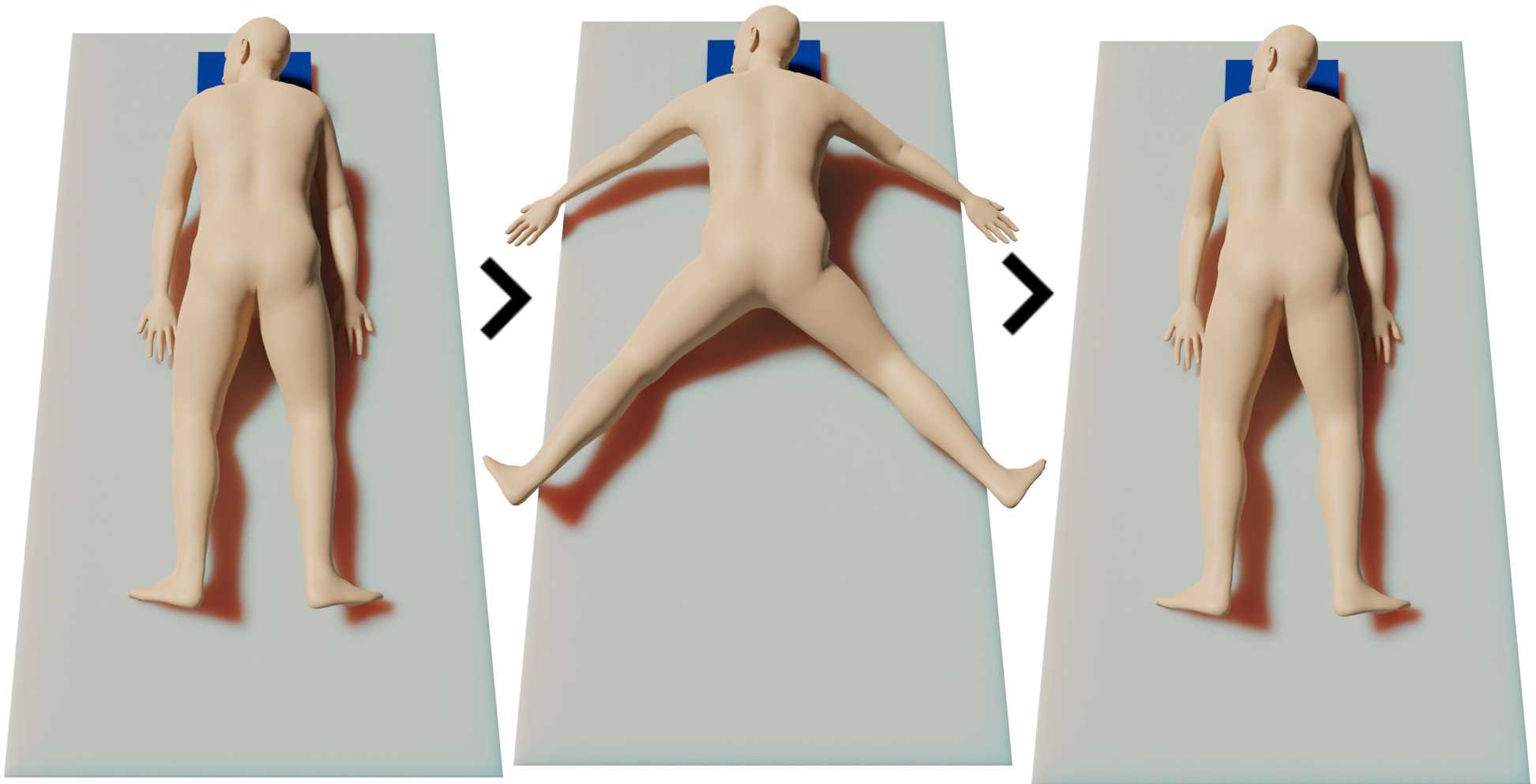}  
  \caption{Belly-down spread}
  \label{fig:sub5}
\end{subfigure}
\begin{subfigure}{.49\linewidth}
  \centering
  \includegraphics[width=.75\linewidth]{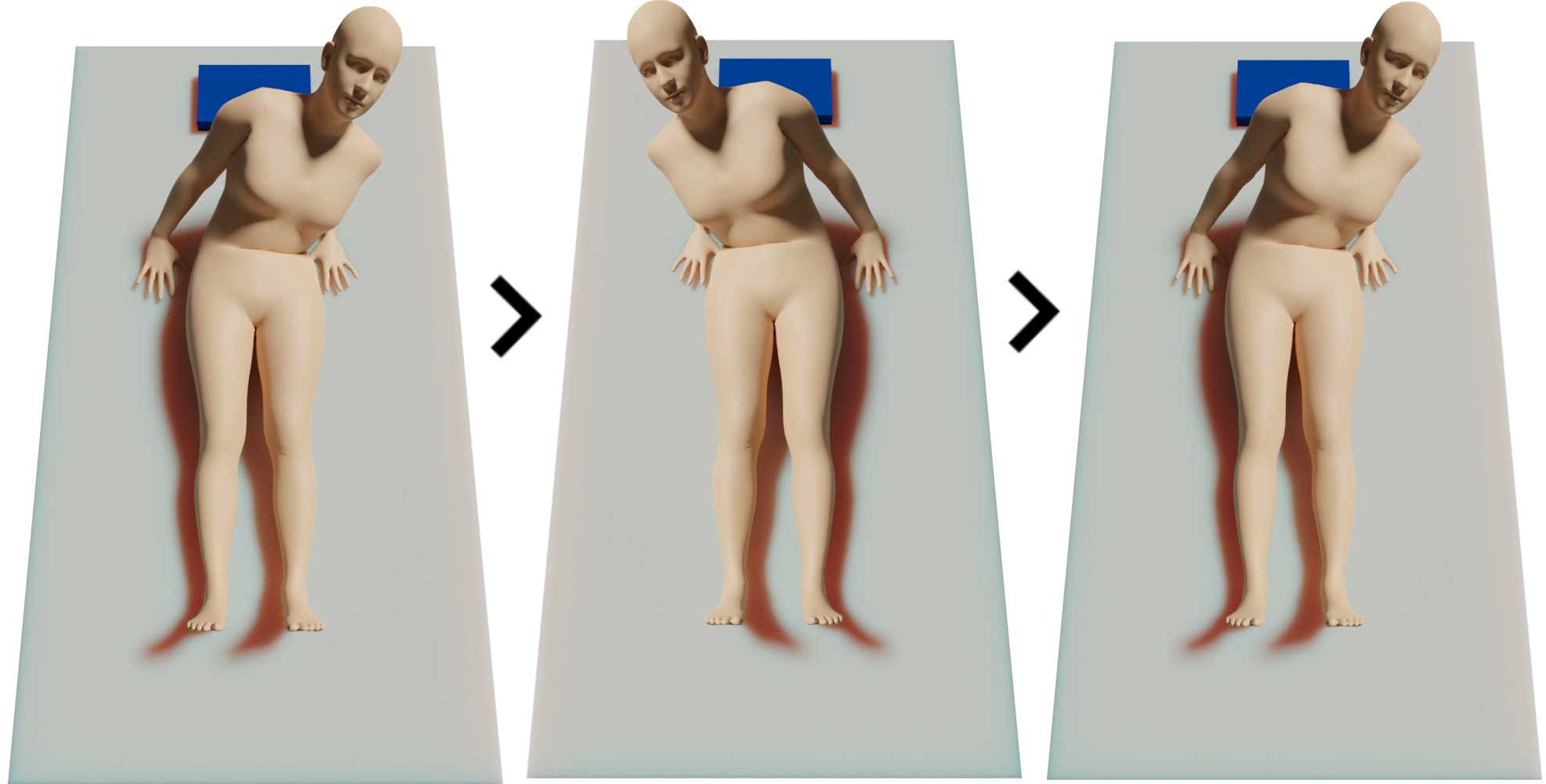}  
  \caption{Torso lean}
  \label{fig:sub6}
\end{subfigure}
\begin{subfigure}{\linewidth}
  \centering
  \includegraphics[width=.75\linewidth]{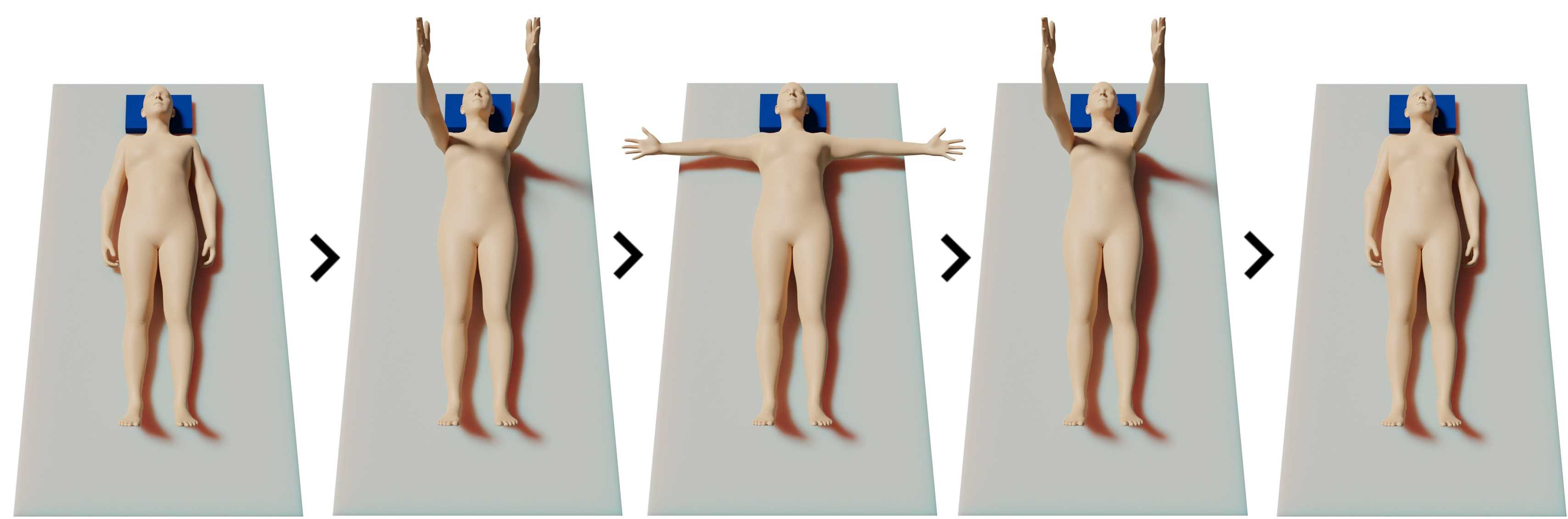}  
  \caption{Stretched arms}
  \label{fig:sub7}
\end{subfigure}
\begin{subfigure}{\linewidth}
  \centering
  \includegraphics[width=.375\linewidth]{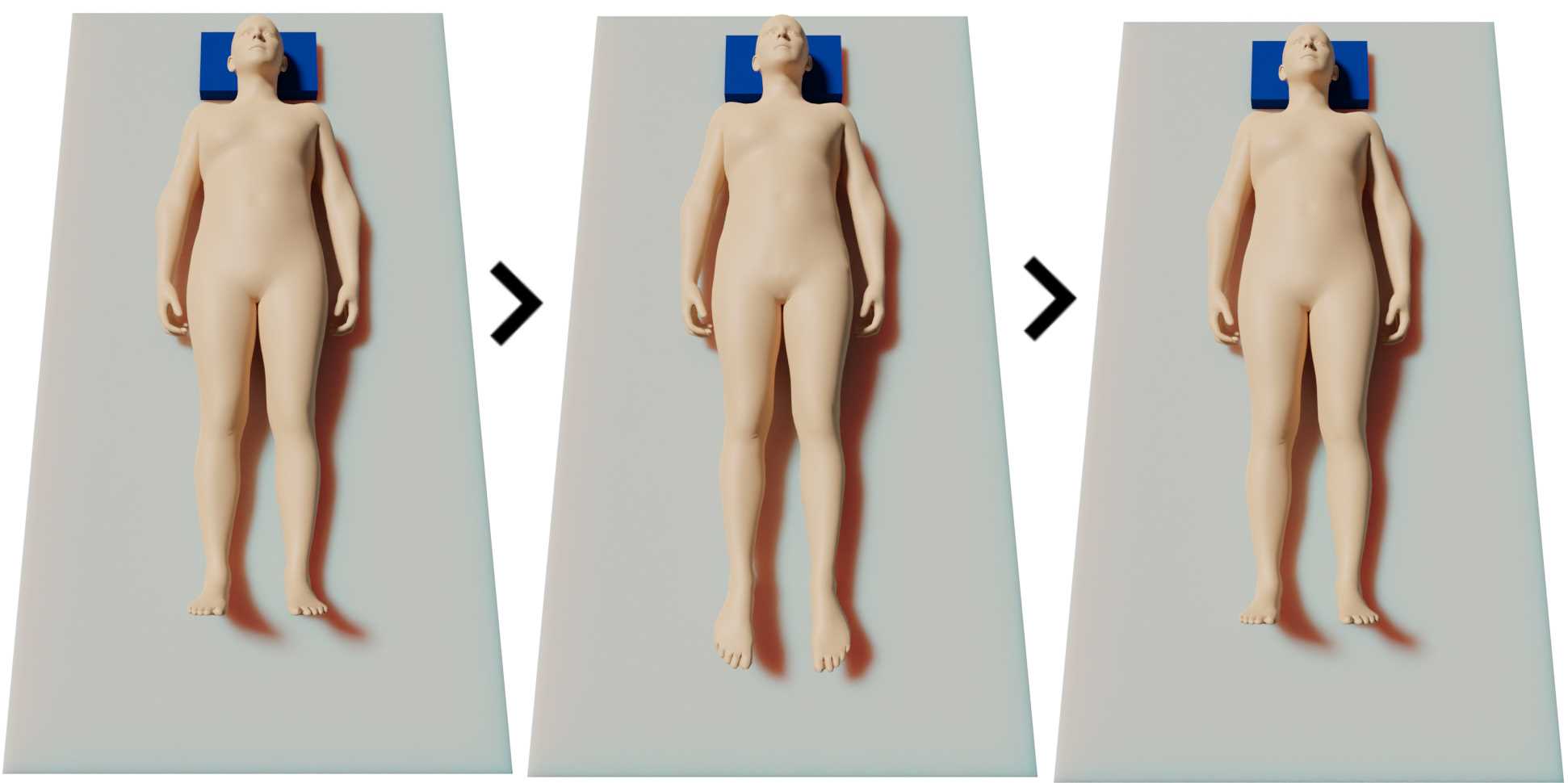}  
  \setlength{\belowcaptionskip}{-10pt}
  \caption{Feet stretched}
  \label{fig:sub8}
\end{subfigure}
\setlength{\belowcaptionskip}{-8pt}
\caption{3D models performing the 8 movement sequences.}
\label{fig:movall}
\end{figure}

After careful consideration, 8 motion sequences were selected, 5 involving the legs and 4 involving the arms (1 involving both the arms and the legs). Each of them was given a code name, as well as a number. They are displayed in figure \ref{fig:movall}.

\begin{enumerate}[topsep=0em, partopsep=0em,itemsep=-0.2em]
    \item \textbf{Foot to knee: }The participant starts in a resting position facing upwards, they then lift their left foot and place it on top of the right knee, then they reset back to the starting position and repeat the previous steps but with the right foot.
    This sequence was chosen because it severely displaces the blanket with every repetition, and it includes some self-occlusion (Fig. \ref{fig:movall}.a).
    \item \textbf{Knee bend: }The participant starts in a resting position facing upwards with their right knee bent and their right foot on the bed next to the left knee, they then switch to the same pose but flipped.
    This sequence was chosen because, while the overall shape of the legs is not visible under the blanket, the deformation of the blanket contains information about the position of the knee (Fig. \ref{fig:movall}.b).
    \item \textbf{Swinging legs: }The participant starts in a resting position on the side, with one knee stretched and the other bent at a 90º angle. They then flip which knee is stretched and which is bent.
The side on which the participants lie down is randomly decided at the time of recording.
    This sequence was chosen because the overall shape of the legs can be distinguished under the blankets, and because it has a lot of self-occlusion (Fig. \ref{fig:movall}.c).
    \item \textbf{Hands to shoulders: }The participant starts in a resting position facing upwards, then they bring their hands up to the opposite shoulders.
    This sequence was chosen because the amount of interaction between the hands and the blanket varies with each repetition (Fig. \ref{fig:movall}.d).
    \item \textbf{Belly-down spread: }The participant starts in a resting position facing down, they then spread their arms and legs.
    This sequence was chosen because it significantly displaces the blanket with each repetition, and because it involves moving both the arms and legs simultaneously without being difficult to coordinate (Fig. \ref{fig:movall}.e).
    \item \textbf{Torso lean: }The participant starts in a leaning position with the right forearm lying down on the bed and the left arm stretched, they then switch to the same pose but flipped. 
    This sequence was chosen because it includes movement of the torso, which is generally static in the other sequences (Fig. \ref{fig:movall}.f).
    \item \textbf{Stretched arms: }The participant starts in a resting position facing up with their arms stretched next to them, they then raise their arms up, followed by bringing them down into a T-pose, then bringing them back up before going back to the resting position.
    This sequence was chosen because it includes a large portion of the range of motion of the shoulders plus it is partially vertical, which is a dimension the other sequences don’t prominently explore (Fig. \ref{fig:movall}.g).
    \item \textbf{Feet stretched: }The participant starts in a resting position facing up with their feet at 90º to the rest of the leg, they then stretch their feet as far as they can.
    This sequence was chosen because it consists only of movements of the feet, which are very difficult to notice under blankets (Fig. \ref{fig:movall}.h). 
\end{enumerate}

\subsection{Dataset Acquired}

A total of 405 videos (303965 frames) were recorded, the distribution of the different labels is described in Table \ref{tab:framecount}. The parameters used in the recordings are described in Table \ref{tab:azureparams}

\begin{table}[h!]
\resizebox{0.49\textwidth}{!}{%
\begin{tabular}{c|@{\hskip 0.2em}c@{\hskip 0.2em}c@{\hskip 0.2em}c@{\hskip 0.2em}c@{\hskip 0.2em}c@{\hskip 0.2em}c@{\hskip 0.2em}c@{\hskip 0.2em}c@{\hskip 0.2em}|@{\hskip 0.2em}c@{\hskip 0.2em}c@{\hskip 0.2em}c@{\hskip 0.2em}|@{\hskip 0.2em}c@{\hskip 0.2em}c@{\hskip 0.2em}c@{\hskip 0.2em}c@{\hskip 0.2em}|@{\hskip 0.2em}c}
\setlength{\tabcolsep}{1pt}
     & \multicolumn{8}{c@{\hskip 0.2em}|@{\hskip 0.2em}}{Sequences} & \multicolumn{3}{c@{\hskip 0.2em}|@{\hskip 0.2em}}{BIPs} & \multicolumn{4}{c@{\hskip 0.2em}|@{\hskip 0.2em}}{Hand Pos.} & \\
     
     Lab. & 1 & 2 & 3 & 4 & 5 & 6 & 7 & 8 & Off & Half & Full & 1 & 2 & 3 & 4 & \textbf{Tot.} \\
     \hline
     Vid. & 54 & 54 & 54 & 54 & 45 & 54 & 45 & 45 & 135 & 135 & 135 & 86 & 117 & 97 & 105 & \textbf{405} \\
     Frs. & 40 & 35 & 44 & 48 & 29 & 39 & 32 & 36 & 98 & 100 & 106 & 58 & 91 & 75 & 79 & \textbf{304}

\end{tabular}}
\setlength{\belowcaptionskip}{-10pt}
\caption{The labels provided in the dataset, and the corresponding number of videos and frames. Lab.- Labels, Vid.- Number of videos, Frs.- Number of frames ($\times10^3$), Tot.- Total}
\label{tab:framecount}
\end{table}

\begin{table}[h!]
\centering
\resizebox{0.35\textwidth}{!}{
\begin{tabular}{l|l}
   Parameter  &  Value \\
   \hline 
   RGB Resolution & 1920x1080 \\
   Framerate & 30 fps \\
   Depth Format & NFOV\_2X2BINNED \\
   IR Resolution & 320x288 \\
   IR FoV & 75ºx65º \\
\end{tabular}}
\setlength{\belowcaptionskip}{-10pt}
\caption{The Azure Kinect parameters used in the recordings. FoV stands for Field of View.}
\label{tab:azureparams}
\end{table}

\subsection{Evaluation of BlanketGen}
BlanketSet was used to evaluate the utility of the pipeline implemented in \cite{blanketgen}, which augments datasets with synthetic blanket occlusions in the task of human pose estimation.

\begin{figure}[h!]
\begin{subfigure}[t]{.3\linewidth}
  \centering
  \includegraphics[width=\linewidth]{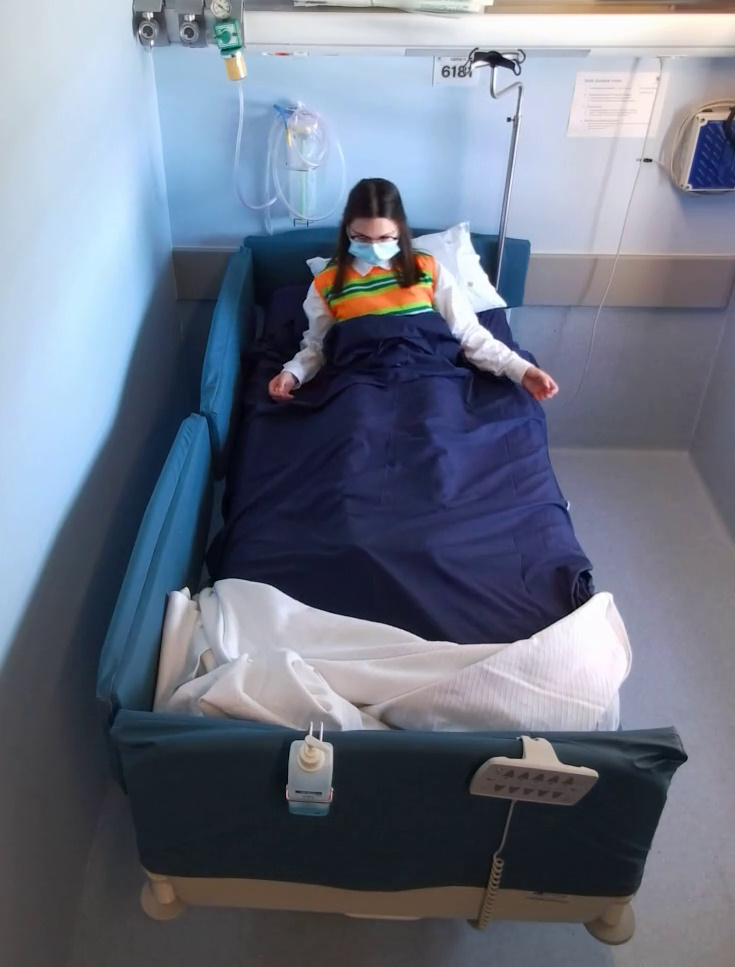}  
  \caption{Original frame}
  \label{fig:shortframe1}
\end{subfigure}
\begin{subfigure}[t]{.3\linewidth}
  \centering
  \includegraphics[width=\linewidth]{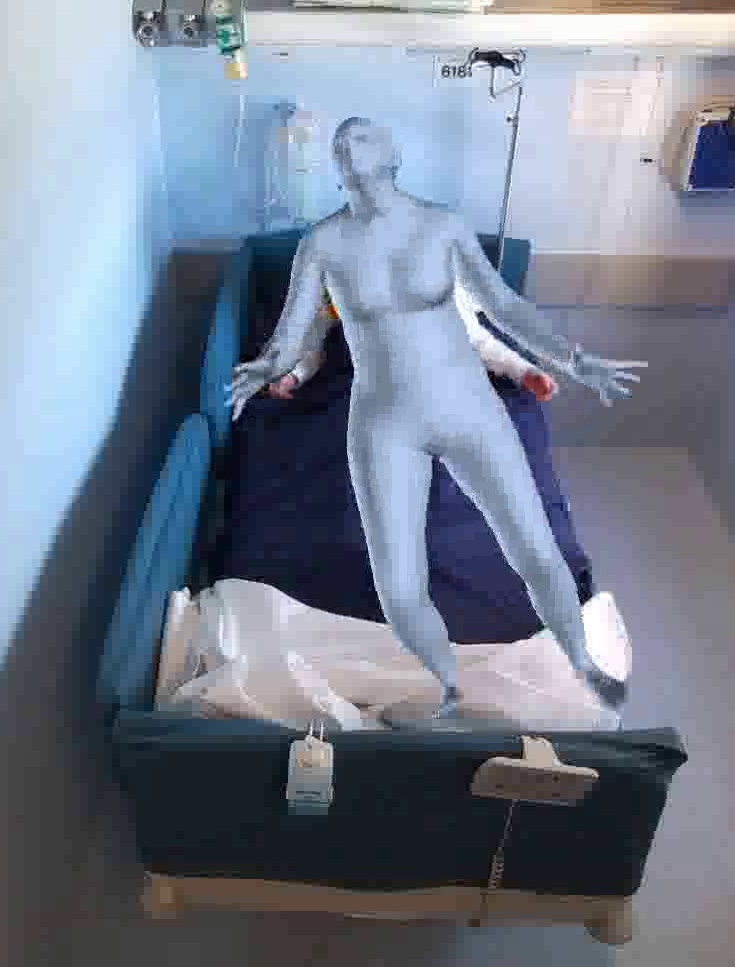}  
  \caption{Pre-trained}
  \label{fig:shortframe2}
\end{subfigure}
\begin{subfigure}[t]{.3\linewidth}
  \centering
  \includegraphics[width=\linewidth]{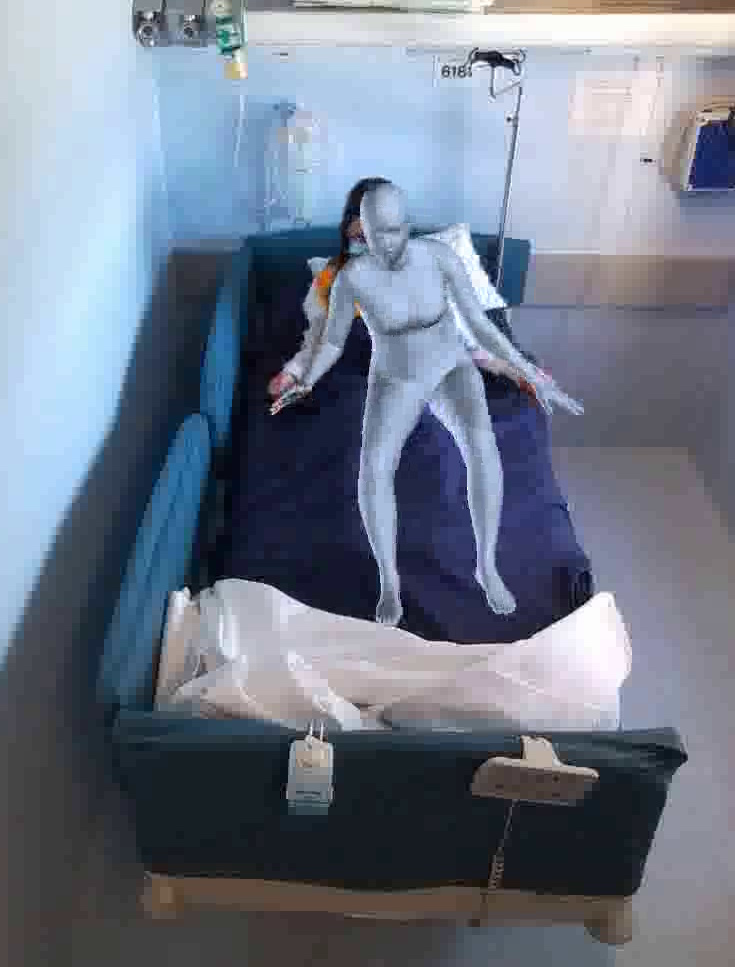}  
  \caption{Fine-tuned}
  \label{fig:shortframe3}
\end{subfigure}
\caption{Example frame from one of the videos selected for evaluation. Two HybrIK models were used to estimate body meshes, the pre-trained one provided by the authors of \cite{hybrik} as well as the one fine-tuned in \cite{blanketgen}.}
\label{fig:qualframe}
\end{figure}

Videos from BlanketSet were processed by both the pre-trained model and the model fine-tuned with synthetic blanket occlusions. For this, one sequence from each participant was selected, with each movement sequence as well as each blanket, being present at least once in this selection. 
The number of videos selected had to be small, so the BIP of the blanket completely off the subject was excluded. BlanketGen had already been evaluated in cases without blanket occlusions in \cite{blanketgen} so evaluating in these cases was not a priority for this paper.
Figure \ref{fig:qualframe} shows a frame from one of the recordings (a), alongside the same frame with body mesh estimation made by the pre-trained model (b), and by the fine-tuned one (c).

A survey was conducted and made public where the annotated videos were rated on a scale from 0 to 10, with 0 being a failure to even locate the subject and 10 being perfect annotations.
This survey was advertised to all students of the Faculty of Engineering of the University of Porto (FEUP) with the dynamic email service provided by the university and it gathered a total of 29 responses.

\section{Results}
Following the methods described previously, an action recognition dataset was acquired with 8 movement sequences being performed in a hospital bed by 14 participants, with 3 levels of blanket occlusion, various blankets, and at different speeds; it also includes recordings of participants with and without masks. 

Videos from BlanketSet were used to evaluate the pipeline implemented in \cite{blanketgen}.
Table \ref{tab:qualitativeselect} describes which recordings were used for the survey as well as the average and standard deviation of the ratings each recording got when annotated with either model, rounded to two decimal cases.

\begin{table}[h!]
    \centering
    \resizebox{\linewidth}{!}{%
    \begin{tabular}{c|ccc|cc}
         BIP & Participant & Seq. & Blanket & Pre-trained\(\uparrow\) & Fine-tuned\(\uparrow\) \\
         \hline
        Half & IT01  & 2 & 1 & \textbf{7.28\(\pm\)1.83} & 6.79\(\pm\)2.06 \\
        Half & IT03 & 1 & 3 & 5.52\(\pm\)1.99 & \textbf{6.07\(\pm\)2.05} \\
        Half & IT05 & 5 & 2 & \textbf{5.10\(\pm\)2.06} & 4.86\(\pm\)1.90 \\
        Half & IT07 & 7 & 1 & 5.62\(\pm\)2.13 & \textbf{6.03\(\pm\)2.20} \\
        Half & IT09 & 1 & 6 & 8.38\(\pm\)1.24 & \textbf{8.52\(\pm\)1.30} \\
        Half & IT11 & 4 & 5 & 8.34\(\pm\)1.20 & \textbf{8.48\(\pm\)1.33} \\
        Half & IT16 & 3 & 4 & \textbf{6.07\(\pm\)}1.62 & 5.59\(\pm\)1.74 \\
        Half & IT20 & 5 & 6 & 8.14\(\pm\)1.68 & \textbf{8.34\(\pm\)1.49} \\
        \hline
        Full & IT02 & 6 & 2 & 3.55\(\pm\)1.84 & \textbf{4.34\(\pm\)1.95} \\
        Full & IT04 & 4 & 1 & 0.28\(\pm\)0.80 & \textbf{0.38\(\pm\)1.08} \\
        Full & IT06 & 8 & 3 & \textbf{0.69\(\pm\)1.61} & 0.45\(\pm\)1.12 \\
        Full & IT08 & 3 & 2 & 1.00\(\pm\)1.63 & \textbf{1.10\(\pm\)1.92} \\
        Full & IT10 & 2 & 1 & 0.55\(\pm\)1.30 & \textbf{0.66\(\pm\)1.63} \\
        Full & IT13 & 6 & 6 & \textbf{5.62\(\pm\)1.70} & 5.34\(\pm\)1.90 \\
        Full & IT17 & 8 & 5 & 0.72\(\pm\)1.16 & \textbf{1.52\(\pm\)}1.50 \\
    \end{tabular}}%
    \setlength{\belowcaptionskip}{-10pt}
    \caption{Recordings selected for evaluation and evaluations of their annotations (Seq.- Sequence). The best result for each sequence is in bold. 
    Note: participants IT03 and IT17 were the same person with and without face-mask.}

    \label{tab:qualitativeselect}
\end{table}

\begin{table}[h!]
    \centering
    \setlength{\belowcaptionskip}{-20pt}
    \resizebox{0.49\textwidth}{!}{
    \begin{tabular}{c|cc|c}
         BIP & Pre-trained\(\uparrow\) & Fine-tuned (ours)\(\uparrow\) & Difference \\
         \hline
         Half & 6.81\(\pm\)2.16 & \textbf{6.84\(\pm\)2.22} & 0.03 \((p < 0.66)\)\\
         Full & 1.77\(\pm\)2.37 & \textbf{1.97\(\pm\)2.47} & \textbf{0.20 \((p < 0.02)\)}\\
         \hline
         Total & 4.46\(\pm\)3.38 & \textbf{4.57\(\pm\)3.37} & \textbf{0.11 \((p < 0.02)\)}
    \end{tabular}}
    \setlength{\belowcaptionskip}{-4pt}
    \caption{The average ratings on a scale of 0 to 10 by BIP. The differences between the results of the two models are listed on the right as well as their p-values rounded up to two decimals.}
    \label{tab:qualshortresults}
\end{table}

As expected, the determining factor for the quality of the annotations was the blanket initial position (BIP); table \ref{tab:qualshortresults} summarizes these results as well as the rating difference between both models and a symmetric 95\% confidence interval for these differences calculated with a two-tailed paired t-test. Fine-tuning led to a significant improvement in performance when full-body blanket occlusions were present (+0.20, \(p<0.02\)), but not when only half the body was occluded (+0.03, \(p<0.66\)).

\section{Discussion}
As an action recognition dataset, BlanketSet can be used to develop architectures to use in epileptic seizure classification. Besides this, since every action was recorded with 3 levels of occlusion, if a two-stage system is implemented to perform action recognition on BlanketSet then the performance of the first stage could be indirectly evaluated by comparing the performance of the action recognition with different levels of blanket occlusion.

By using BlanketSet in conjunction with BlanketGen \cite{blanketgen}, such a two-stage approach could be possible where the first stage estimates the poses of the subjects. If this approach proves successful then it could be used for the automatic classification of epileptic seizures.

The original paper on BlanketGen \cite{blanketgen} found that the data augmentation improved the performance when synthetic occlusions were present, but it couldn't explore whether the improvement extended to the real world. 

BlanketSet helped bridge this gap and, in this paper, results were obtained which evaluate BlanketGen in the real world.

These results are consistent with what was found in the BlanketGen paper, namely that training on the augmented dataset improved the performance when blanket occlusions were present and that the bigger the occlusions were the greater the improvement.

\section{Conclusion}
In this paper, we introduce BlanketSet, an action recognition dataset that has the potential to aid in the research of human motion analysis in clinical contexts where the patients are lying down in bed, particularly in the cases of epilepsy monitoring and sleep analysis.
As an example of a use for BlanketSet, it was utilized to evaluate the performance of BlanketGen \cite{blanketgen} in the task of human pose estimation.

\section{Acknowledgements}
We would like to thank the University Hospital Center of São João, particularly Dr. Ricardo Rego and the EMU staff, for letting us use their facilities to record BlanketSet.

\bibliographystyle{IEEEtran}
\bibliography{main}

\end{document}